\documentclass[conference, 10pt]{IEEEtran}
\IEEEoverridecommandlockouts 
\usepackage[]{graphicx}
\usepackage{caption}
\usepackage{subfig} 
\usepackage{amsmath}
\usepackage{amssymb}
\usepackage{epsfig}
\usepackage{cite}
\usepackage{color}
\usepackage{balance}
\usepackage{amsmath}
\usepackage{bm}
\usepackage{amsfonts} 
\usepackage{graphicx}
\usepackage{pdfpages}
\usepackage[T1]{fontenc} 
\usepackage{array}
\usepackage{url}

\usepackage{color}
\usepackage{wrapfig}
\usepackage{lipsum}
\usepackage[hidelinks]{hyperref}
\usepackage{multirow}
\usepackage{tabularx}
\usepackage{tikz}
\usepackage{nth} 
\usepackage{algpseudocode} 
\usepackage{algorithm,algpseudocode}
\usepackage{breqn}
\usepackage{todonotes}
\usepackage{verbatim}

\begin{document}
\title{Detecting Blurred Ground-based Sky/Cloud Images}
\author{\IEEEauthorblockN{
Mayank~Jain\IEEEauthorrefmark{1}\IEEEauthorrefmark{2},
Navya~Jain\IEEEauthorrefmark{3}, 
Yee~Hui~Lee\IEEEauthorrefmark{4}, 
Stefan~Winkler\IEEEauthorrefmark{5}, and
Soumyabrata~Dev\IEEEauthorrefmark{1}\IEEEauthorrefmark{2}
}
\IEEEauthorblockA{\IEEEauthorrefmark{1} The ADAPT SFI Research Centre, Dublin, Ireland}
\IEEEauthorblockA{\IEEEauthorrefmark{2} School of Computer Science, University College Dublin, Ireland}
\IEEEauthorblockA{\IEEEauthorrefmark{3} Ram Lal Anand College, University of Delhi, India}
\IEEEauthorblockA{\IEEEauthorrefmark{4} School of Electrical and Electronic Engineering, Nanyang Technological University, Singapore}
\IEEEauthorblockA{\IEEEauthorrefmark{5} School of Computing, National University of Singapore}

\thanks{This research is partially funded by a research grant from Kaggle, a subsidiary of Google LLC. The ADAPT Centre for Digital Content Technology is funded under the SFI Research Centres Programme (Grant 13/RC/2106\_P2) and is co-funded under the European Regional Development Fund.}
\vspace{-0.6cm}
}

\maketitle

\begin{abstract}
Ground-based whole sky imagers (WSIs) are being used by researchers in various fields to study the  atmospheric events. 
These ground-based sky cameras capture visible-light images of the sky  at regular intervals of time. Owing to the atmospheric interference and camera sensor noise, the captured images often exhibit noise and blur. This may pose a problem in subsequent image processing stages. Therefore, it is important to accurately identify the blurred images. This is a difficult task, as 
clouds have varying shapes, textures, and soft edges whereas the sky acts as a homogeneous and uniform background. 
In this paper, we propose an efficient framework that can identify the blurred sky/cloud images. Using a static external marker, our proposed methodology has a detection accuracy of 94\%. To the best of our knowledge, our approach is the first of its kind in the automatic identification of blurred images for ground-based sky/cloud images. 

\end{abstract}

\IEEEpeerreviewmaketitle

\section{Introduction}

The observation of clouds has an important role in the area of weather forecast, meteorology research, and solar energy forecasting. Moreover, nowadays, ground-based observation devices are preferred over their satellite counterparts~\cite{ye2019supervised}. These whole sky imagers (WSIs) provide images with high temporal and spatial resolution at a low cost~\cite{jain2021wsi}. 
These images provide details of the cloud base, 
unlike satellite images that show the top of the clouds. This is of significance for local weather research or circumsolar disk detection. 

These WSIs provide sky/cloud images with high spatial and temporal resolution. However, these images are also prone to high noise and blur. This may arise from atmospheric dust or water droplets that get accumulated onto the camera lens. Quite often, the sun flare in the captured images causes blur. 
Such blurry images may lead to false detection of cloud pixels in sky/cloud image segmentation~\cite{ICIP1_2014} and image classification~\cite{ICIP2015b} tasks. Therefore, it is important to detect the blurred images 
before proceeding with further imaging analysis. The images captured by these WSIs consist primarily of sky and clouds. The sky behaves as a homogeneous and uniform background, whereas clouds have varying shapes, textures and fuzzy edges. This renders the task of detecting blurred images from a repository of captured images extremely difficult. 

In this paper\footnote{In the spirit of reproducible research, the code related to this paper is available from  \url{https://github.com/navyajain16/Blur-Detection-in-WSI-Images}.}, we propose a framework to accurately identify blurred sky/cloud images with the aid of a static external marker. Such automated framework can easily index the repository of blurred images into \textit{blurred} and \textit{non-blurred} categories. Especially tuned deblurring techniques can then be applied only on the blurred images to enhance the information content in the dataset.

\section{Proposed Methodology for Blur Detection}

Our methodology of automatically identifying blurred images is based on the fundamental paradigm -- it is easier to identify a \textit{blurred regular image} as compared to \textit{blurred sky/cloud image}. The clouds have fuzzy boundaries and sky has a homogeneous background, resulting in a inherent blurry sky/cloud image path. However, if the captured image contains an external marker possessing sharp and distinct boundaries, it is easier to identify the blur around the external marker.

\begin{figure}[htb]
    \centering
    \includegraphics[height=0.2\textwidth]{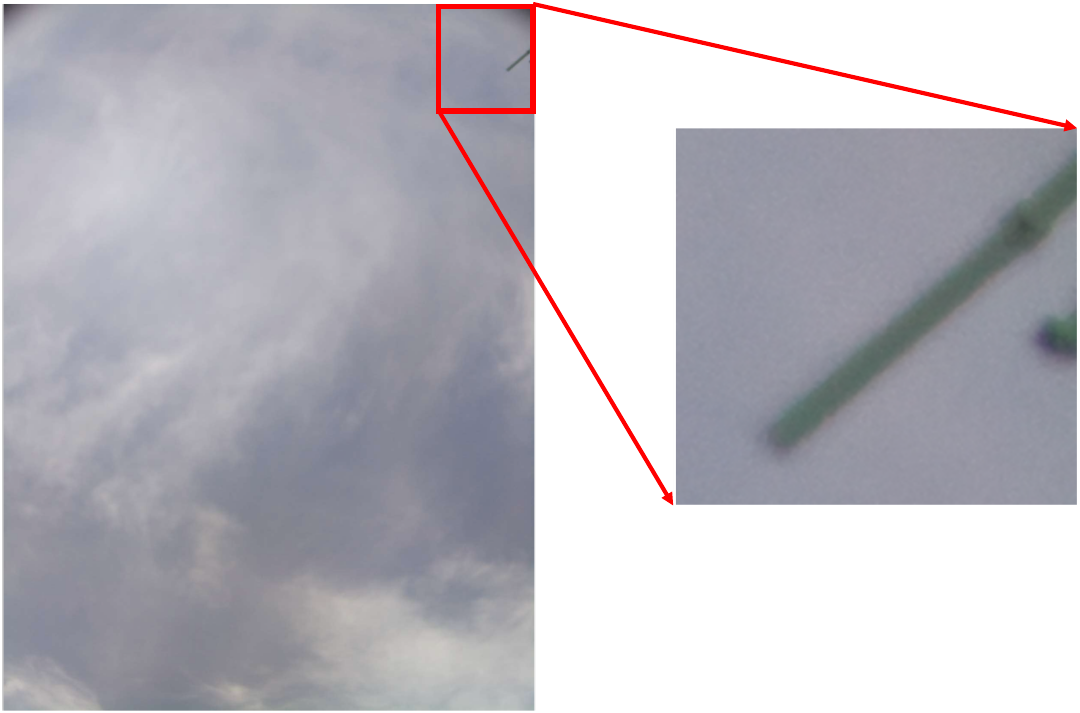}
    \caption{Marker used in our experiments. The second image is the zoomed-in version of the first  showing the marker.}
    \label{fig:Marker}
    \vspace{-0.4cm}
\end{figure}


\begin{figure}[htb]
\centering
\subfloat[\centering Blurred images]{
\includegraphics[width=0.13\textwidth]{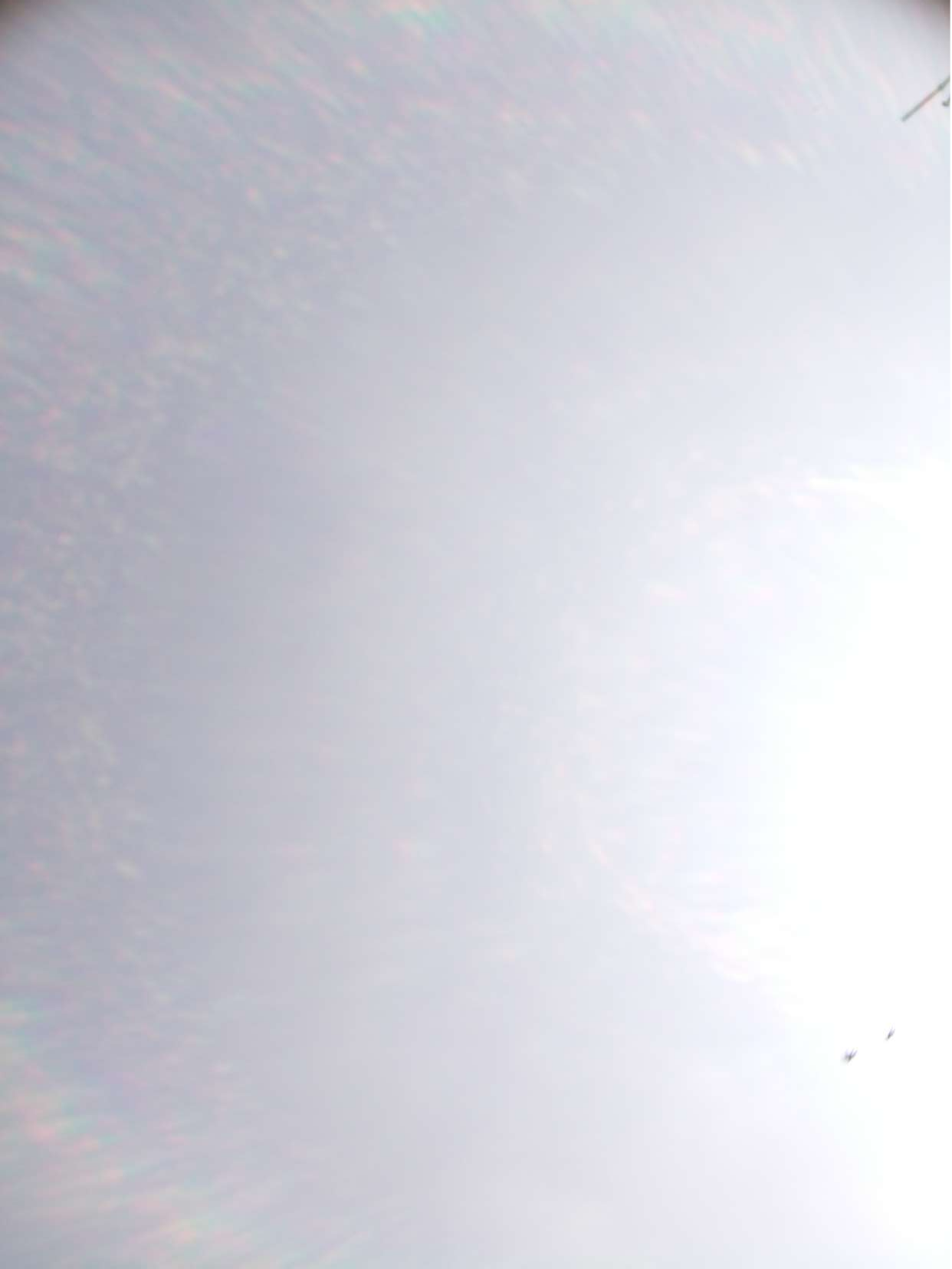}
\includegraphics[width=0.13\textwidth]{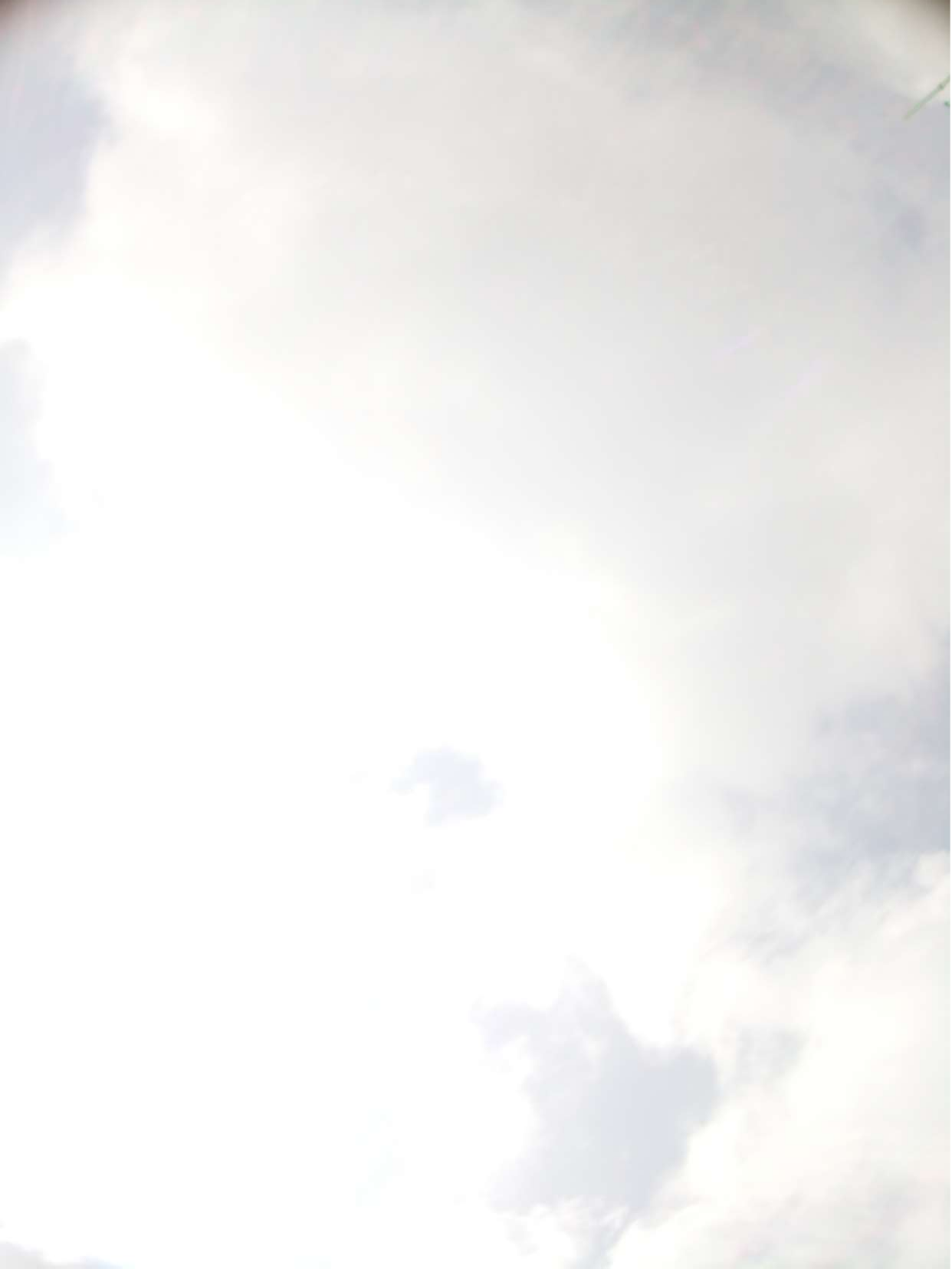}
\includegraphics[width=0.13\textwidth]{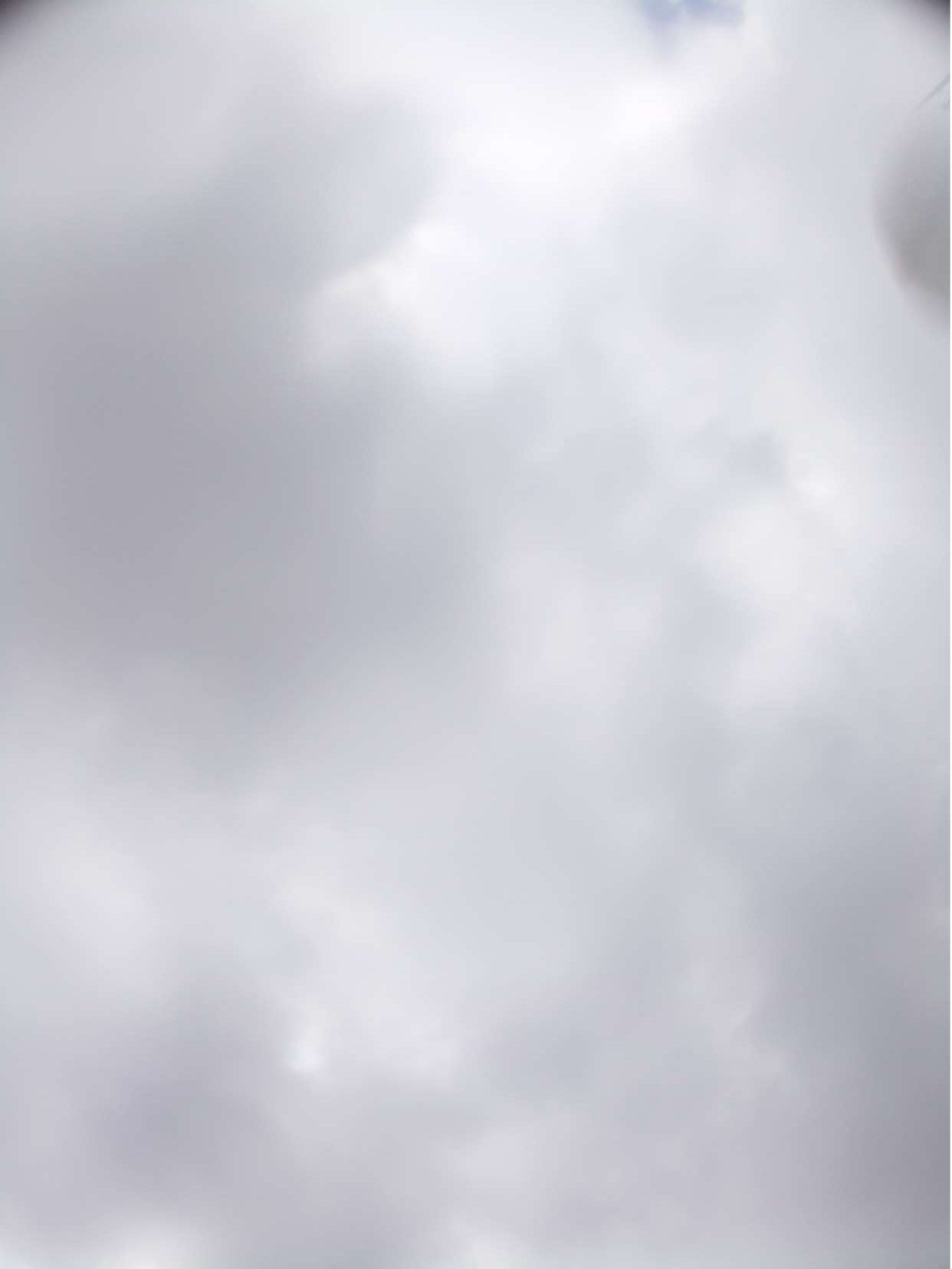}}\\
\subfloat[\centering Non-blurred images]{
\includegraphics[width=0.13\textwidth]{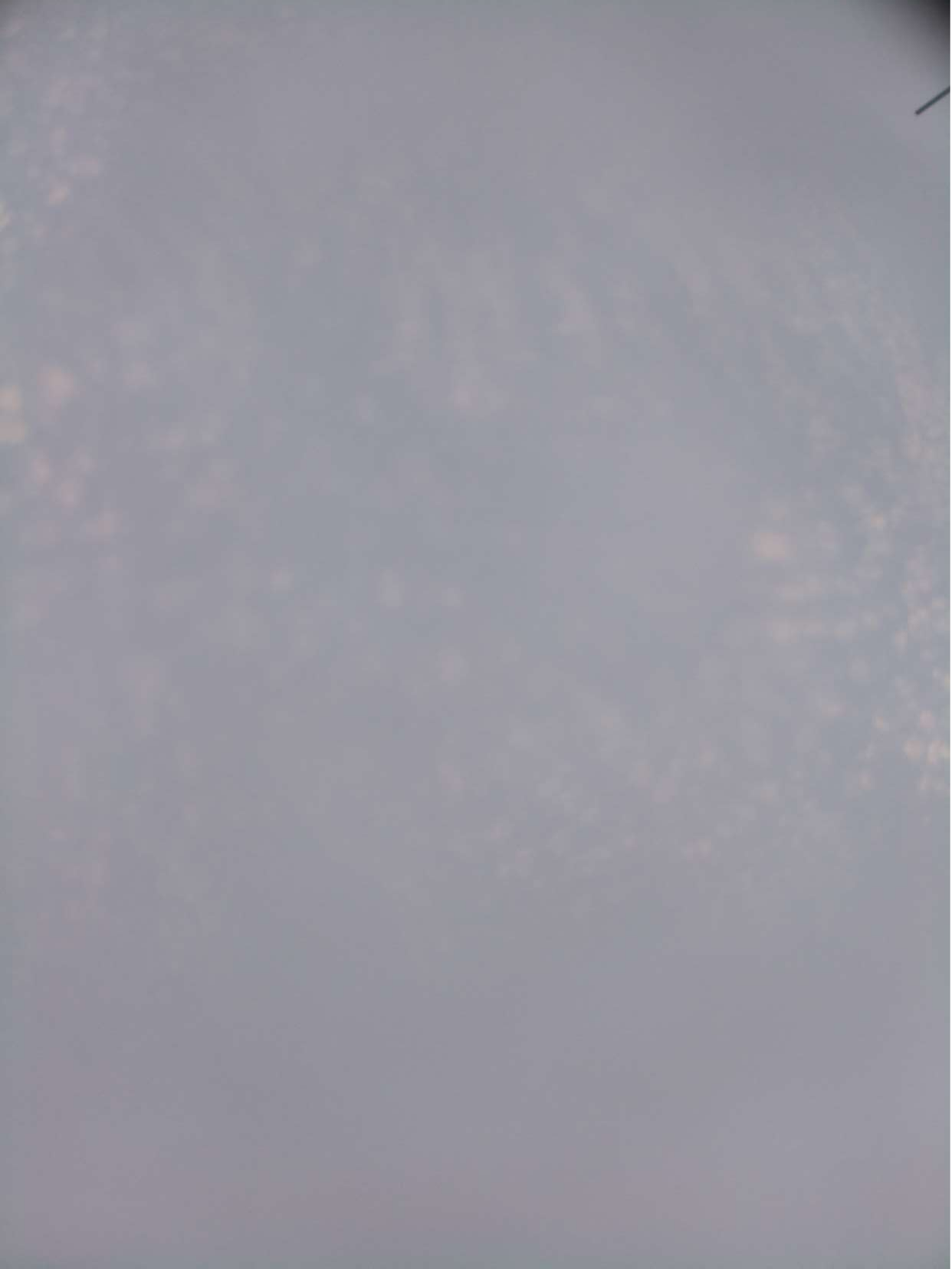}
\includegraphics[width=0.13\textwidth]{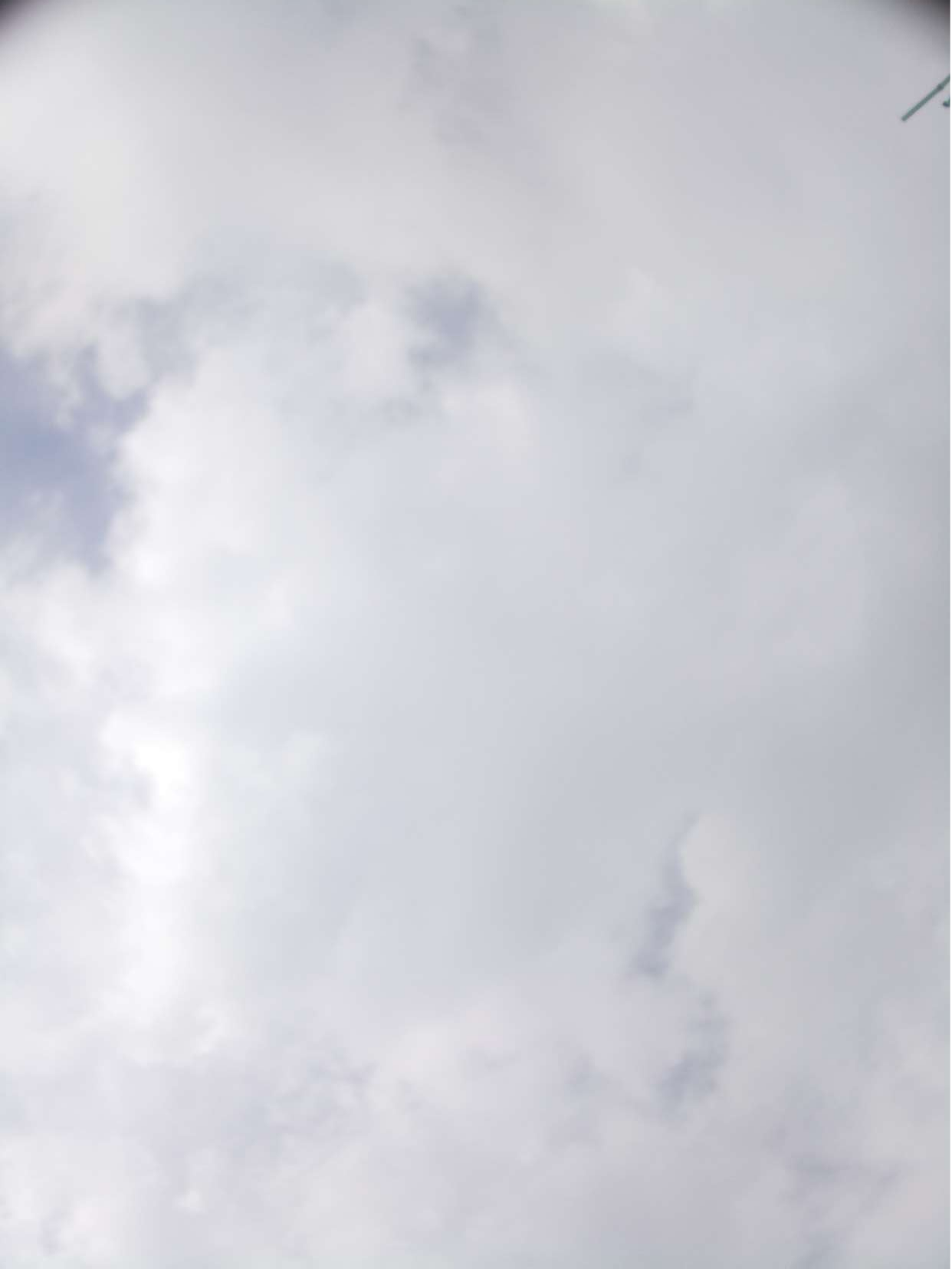}
\includegraphics[width=0.13\textwidth]{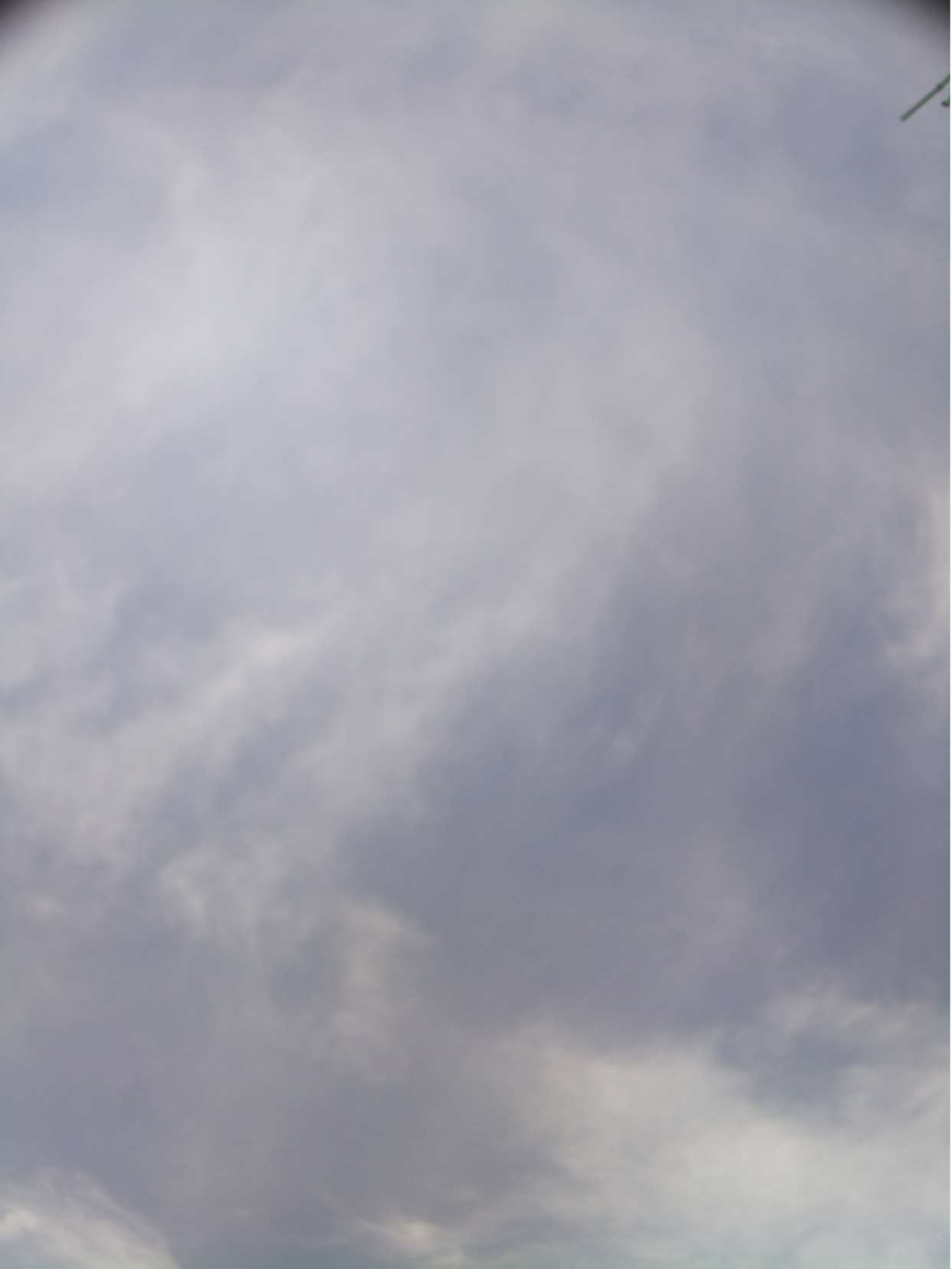}}%
\caption{We show representative examples of blurred and non-blurred cloud images captured by our sky camera. 
\vspace{-0.5cm}
}
\label{fig:blurNoBlurExamples}
\end{figure}

We propose the following to identify blurred images: (a) add an external static marker in the field of view of captured images, (b) analyse the whole- and cropped- version (with marker) of the captured image, (c) detect if the external static marker is blurred or not using a blur-detection metric. If the cropped area around the external marker is blurred, we categorise the image as blurred. Figure~\ref{fig:Marker} illustrates our selected external marker. We chose the pole as an external marker, because it possesses sharp boundaries and remains static throughout all the captured images. We need to ensure that the marker should be a stationary object that does not change its position with time. This will ensure that the blur detection metric is correctly applied to the cropped version of the image containing the marker. 

We use the Laplacian operator~\cite{bansal2016blur} and Fast Fourier Transform (FFT) operator~\cite{pertuz2013analysis} to compute the blur detection metric. These metrics are defined in the following subsections.

\subsection{Laplacian Operator}
Laplacian operator is often used to detect edges in pictures. It highlights regions of an image that contain rapid intensity changes. It is a derivative operator. Unlike other operators like Sobel and Kirsch that measure the first-order derivative, Laplacian measures the second-order derivative of an image. This operator is further classified into a Negative Laplacian operator and a Positive Laplacian operator~\cite{bansal2016blur}. 

\subsection{Fast Fourier Transform (FFT)}
Fast Fourier transform (FFT) calculates the frequencies at different points in the image. This algorithm is based on the set level of frequencies, that decides whether the image is blurry or not. If the amount of high frequency in the image is low, then the image is declared to be blurred. The the limits of the frequency range are determined by the user~\cite{pertuz2013analysis}.

These metrics require a threshold to categorise the image as 
blurred or non-blurred. This selection of the threshold is critical and done by the user~\cite{bansal2016blur}.  A very low threshold will always categorise an image as blurred; whereas a high value will always categorise it as non-blurred. We have manually selected the value of these thresholds in our experiments$:$ $+12$ for the Laplacian method and $-4$ for the FFT method. 


\section{Results \& Discussions}
\subsection{Experimental setup}
The images in our experiments are captured by a ground-based sky camera situated in New Delhi, India~\cite{jain2021wsi}. This camera captures images at regular $5$ minute intervals throughout the day. However, we have considered only day-time images in our study. 
Fig.~\ref{fig:blurNoBlurExamples} shows a few representative images. 
We observe that both blurred and non-blurred images look similar to each other, unless they are examined carefully. Therefore, it becomes extremely difficult to detect blurred patches in these captured images. The presence of an external marker will ease the task of identifying whether the cropped image is blurred or not (\textit{cf.} Fig.\ref{fig:Marker}).


\subsection{Results}
We evaluate the performance of our framework using both Laplacian operator and Fast Fourier Transform operator. We compute the blur metric for both entire image and cropped version of the image containing the marker. Using this computed blur metric, we classify our image as \textit{blurred} and \textit{non-blurred}. We use an image repository of $100$ sky/cloud images and manually label each image as \textit{blurred} or \textit{non-blurred}. These labels act as the ground-truth in our classification framework. Table~\ref{tab:BlurDetectionResults} demonstrates the results on our image repository. 
We observe that the accuracy of detection of blur increases significantly for the cropped-image of the pole for both the algorithms. This makes sense as the presence of an inconspicuous marker can be useful in detecting blurred 
sky/cloud images. Furthermore, the Laplacian operator performs better than the FFT operator. Therefore, we propose the use of Laplacian operator on the cropped version of the image (containing marker) to identify the blurred images.



\begin{table}[!ht]
    \centering
    \caption{Blur detection accuracy for Laplacian and FFT methods for complete image vs only marker portion of the image. Better accuracies are depicted in boldface.}
    \label{tab:BlurDetectionResults}
    \begin{tabular}{c c c}
        \hline
        Image Analyzed & Accuracy (Laplacian) & Accuracy (FFT)\\
        \hline
        Complete Image & $64\%$ & $61\%$ \\
        Cropped Image w/Marker & $\bm{94\%}$ & $\bm{92\%}$ \\
        \hline
    \end{tabular}
\end{table}

\section{Conclusion \& Future Work}
In this paper, we proposed a technique of automatically identifying blurred images in an image repository. Using an external static marker, we obtained good accuracy in identifying the blurred images. In the future, we intend to extend this approach to nighttime images, and use other external markers.

\vspace{-0.1cm}

\bibliographystyle{IEEEtran.bst}

\end{document}